\documentclass[runningheads,a4paper]{llncs}[2015/06/24]

\usepackage{cmap}
\usepackage[T1]{fontenc}

\usepackage{graphicx}

\usepackage[ngerman,english]{babel}
\addto\extrasenglish{\languageshorthands{ngerman}\useshorthands{"}}

\usepackage[%
rm={oldstyle=false,proportional=true},%
sf={oldstyle=false,proportional=true},%
tt={oldstyle=false,proportional=true,variable=true},%
qt=false%
]{cfr-lm}
\usepackage[math]{blindtext}

\usepackage{cite}

\usepackage{paralist}

\usepackage{csquotes}

\usepackage{microtype}

\usepackage{subcaption}

\usepackage{url}
\makeatletter
\g@addto@macro{\UrlBreaks}{\UrlOrds}
\makeatother

\usepackage{xcolor}

\usepackage{pdfcomment}
\hypersetup{hidelinks,
   colorlinks=true,
   allcolors=black,
   pdfstartview=Fit,
   breaklinks=true}
\usepackage[all]{hypcap}

\usepackage[capitalise,nameinlink]{cleveref}
\crefname{section}{Sect.}{Sect.}
\Crefname{section}{Section}{Sections}

\usepackage{xspace}

\usepackage{bm}
\usepackage{amsmath}
\usepackage{mathtools}

\DeclareFontFamily{U}{MnSymbolC}{}
\DeclareSymbolFont{MnSyC}{U}{MnSymbolC}{m}{n}
\DeclareFontShape{U}{MnSymbolC}{m}{n}{
    <-6>  MnSymbolC5
   <6-7>  MnSymbolC6
   <7-8>  MnSymbolC7
   <8-9>  MnSymbolC8
   <9-10> MnSymbolC9
  <10-12> MnSymbolC10
  <12->   MnSymbolC12%
}{}
\DeclareMathSymbol{\powerset}{\mathord}{MnSyC}{180}

\begin{document}

\title{Automatic Instrument Segmentation in Robot-Assisted Surgery Using Deep Learning}
\titlerunning{Deep Learning for Surgical Instrument Segmentation}



\author{Alexey A. Shvets\inst{1}, Alexander Rakhlin\inst{2}, Alexandr A. Kalinin\inst{3}, \\ \and Vladimir I. Iglovikov\inst{4}}
 \authorrunning{Shvets et al.}

 \institute{
  Massachusetts Institute of Technology, Cambridge, MA 02142, USA\\ \email{shvets@mit.edu}
  \and Neuromation OU, Tallinn, 10111 Estonia\\ \email{rakhlin@neuromation.io} 
  \and University of Michigan, Ann Arbor, MI 48109, USA\\ \email{akalinin@umich.edu}
  \and Lyft Inc., San Francisco, CA 94107, USA\\ \email{iglovikov@gmail.com}  
  }

\maketitle

\begin{abstract}
Semantic segmentation of robotic instruments is an important problem for the robot-assisted surgery. One of the main challenges is to correctly detect an instrument's position for the tracking and pose estimation in the vicinity of surgical scenes. Accurate pixel-wise instrument segmentation is needed to address this challenge. In this paper we describe our deep learning-based approach for robotic instrument segmentation. Our approach demonstrates an improvement over the state-of-the-art results using several novel deep neural network architectures. It addressed the binary segmentation problem, where every pixel in an image is labeled as an instrument or background from the surgery video feed. In addition, we solve a multi-class segmentation problem, in which we distinguish between different instruments or different parts of an instrument from the background. In this setting, our approach outperforms other methods for automatic instrument segmentation thereby providing state-of-the-art results for these problems.
The source code for our solution is made publicly available.
\end{abstract}


\section{Introduction}\label{sec:intro}

Research in robotics promises to revolutionize surgery towards safer, more consistent and minimally invasive intervention \cite{burgner2015continuum, munzer2018content}. New developments continues on the way to robot-assisted systems and moving toward a future with fully autonomous robotic surgeons. Thus far, the most widespread surgical system is the da Vinci robot, which has already proved its favor via remote controlled laparoscopic surgery in gynecology, urology, and general surgery \cite{burgner2015continuum}.


Information in a surgical console of a robot-assisted surgical system includes valuable details for intra-operative guidance that can help the decision making process. This information is usually represented as 2D images or videos that contain surgical instruments and patient tissues. Understanding these data is a complex problem that involves the tracking and pose estimation for surgical instruments in the vicinity of surgical scenes. A critical component of this process is semantic segmentation of the instruments in the surgical console. Semantic segmentation of robotic instruments is a difficult task by the virtue of light changes such as shadows and specular reflections, visual occlusions such as blood and camera lens fogging, and due to the complex and dynamic nature of background tissues. Segmentation masks can be used to provide a reliable input to instrument tracking systems. Therefore, there is a compelling need for the development of accurate and robust computer vision methods for semantic segmentation of surgical instruments from operational images and video. 

There is a number of vision-based methods developed for the robotic instrument detection and tracking \cite{munzer2018content}. Instrument-background segmentation can be treated as a binary or instance segmentation problem for which classical machine learning algorithms have been applied using color and/or texture features  \cite{speidel2006tracking, doignon2007segmentation}. Later applications addressed this problem as semantic segmentation, aiming to distinguish between different instruments or their parts \cite{pezzementi2009articulated, bouget2015detecting}.

Recently, deep learning-based approaches demonstrated performance improvements over conventional machine learning methods for many problems in biomedicine \cite{ching2017opportunities, kalinin2018deep}. In the domain of medical imaging, convolutional neural networks (CNN) have been successfully used, for example, for breast cancer histology image analysis \cite{rakhlin2018deep}, bone disease prediction \cite{tiulpin2018automatic} and age assessment \cite{iglovikov2017pediatric}, and other problems \cite{ching2017opportunities}. Previous deep learning-based applications to robotic instrument segmentation have demonstrated competitive performance in binary segmentation \cite{garcia2017toolnet, attia2017surgical} and promising results in multi-class segmentation \cite{pakhomov2017deep}.

In this paper, we present a deep learning-based solution for robotic instrument semantic segmentation that achieves state-of-the-art results in both binary and multi-class setting. We used this method to produce a submission to the MICCAI 2017 Endoscopic Vision SubChallenge: Robotic Instrument Segmentation \cite{miccai2017} to achieve one of the top results. Here we describe the details of the solution based on a modification of the U-Net model \cite{ronneberger2015u, iglovikov2017satellite}. Moreover, we provide further improvements over this solution utilizing recent deep architectures: TernausNet \cite{iglovikov2018ternausnet} and a modified LinkNet \cite{chaurasia2017linknet}.

\begin{figure}[!t]
\centering
\includegraphics[width=\linewidth]{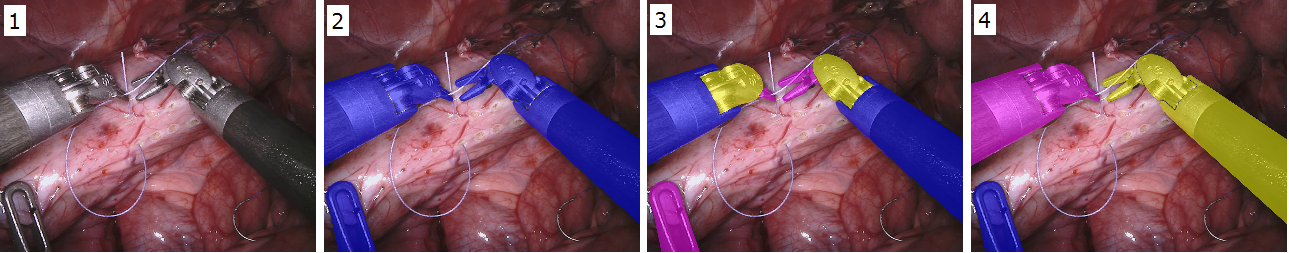}
\caption{A snapshot from a robotic surgical video that contains robotic instruments and patient tissues: (1) original video frame; (2) binary segmentation of robotic instruments shown in blue and tissue that serves as a background; (3) multi-class segmentation of robotic instruments where each class corresponds to a different part of the robotic instrument (3 classes: rigid shaft, articulated wrist and claspers); and (4) multi-class segmentation of robotic instruments where each class corresponds to a different robotic instrument (7 classes).}
\label{fig::classes}
\end{figure}

\section{Methods}
\subsection{Dataset}
The training dataset consists of $8\times225$-frame sequences of high resolution stereo camera images acquired from a da Vinci Xi surgical system during several different porcine procedures, see \cite{miccai2017}. Training sequences are provided with 2 Hz frame rate to avoid redundancy. Every video sequence consists of two stereo channels taken from left and right cameras and has a $1920\times1080$ pixel resolution in RGB format. To remove black canvas and extract original $1280\times1024$ camera images from the frames, an image has to be cropped starting from the pixel at the $(320, 28)$ position. Ground truth labels are provided for left frames only, therefore only left channel images are used for training. The articulated parts of the robotic surgical instruments, such as a rigid shaft, an articulated wrist and claspers have been hand labelled in each frame. Ground truth labels are encoded with numerical values $(10, 20, 30, 40, 0)$ and assigned to each part of an instrument or background. Furthermore, there are instrument type labels that categorize instruments in following categories: left/right prograsp forceps, monopolar curved scissors, large needle driver, and a miscellaneous category for any other surgical instruments.

The test dataset consists of $8\times75$-frame sequences containing footage sampled immediately after each training sequence and 2 full 300-frame sequences, sampled at the same rate as the training set. Under the terms of the challenge, participants should exclude the corresponding training set when evaluating on one of the 75-frame sequences.


\subsection{Network architectures}
In this work we evaluate 4 different deep architectures for segmentation: U-Net \cite{ronneberger2015u, iglovikov2017satellite}, 2 modifications of TernausNet \cite{iglovikov2018ternausnet}, and a modification of LinkNet \cite{chaurasia2017linknet}.

\begin{figure*}[!b]
\includegraphics[width=\linewidth]{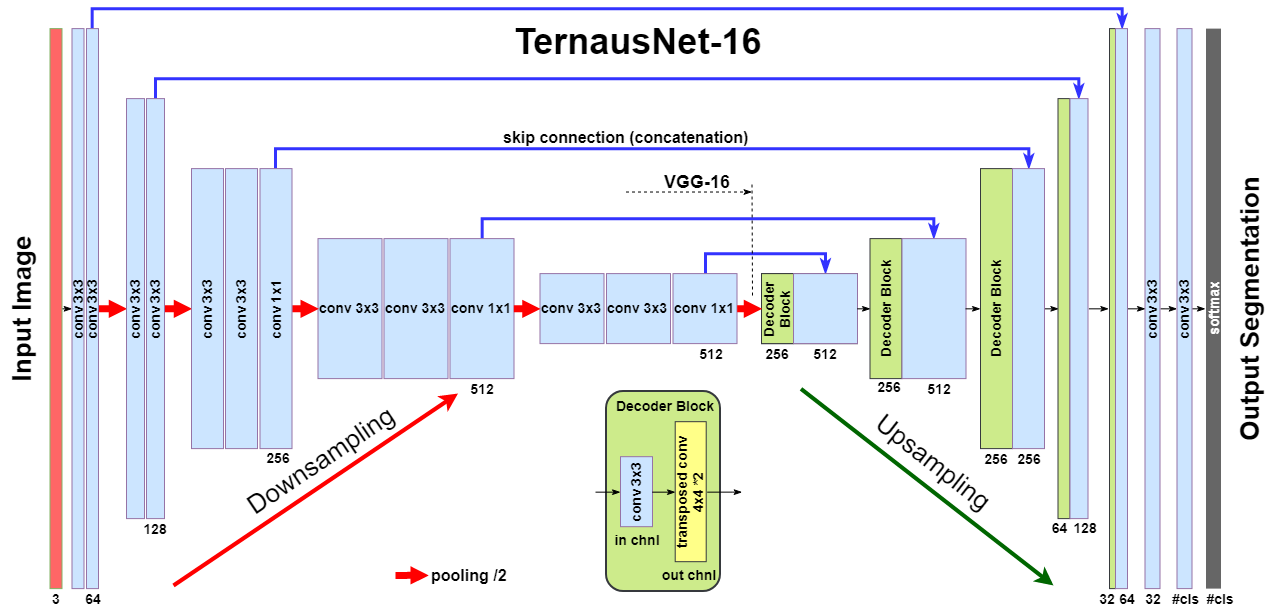}
\includegraphics[width=\textwidth]{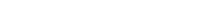}
\includegraphics[width=\linewidth]{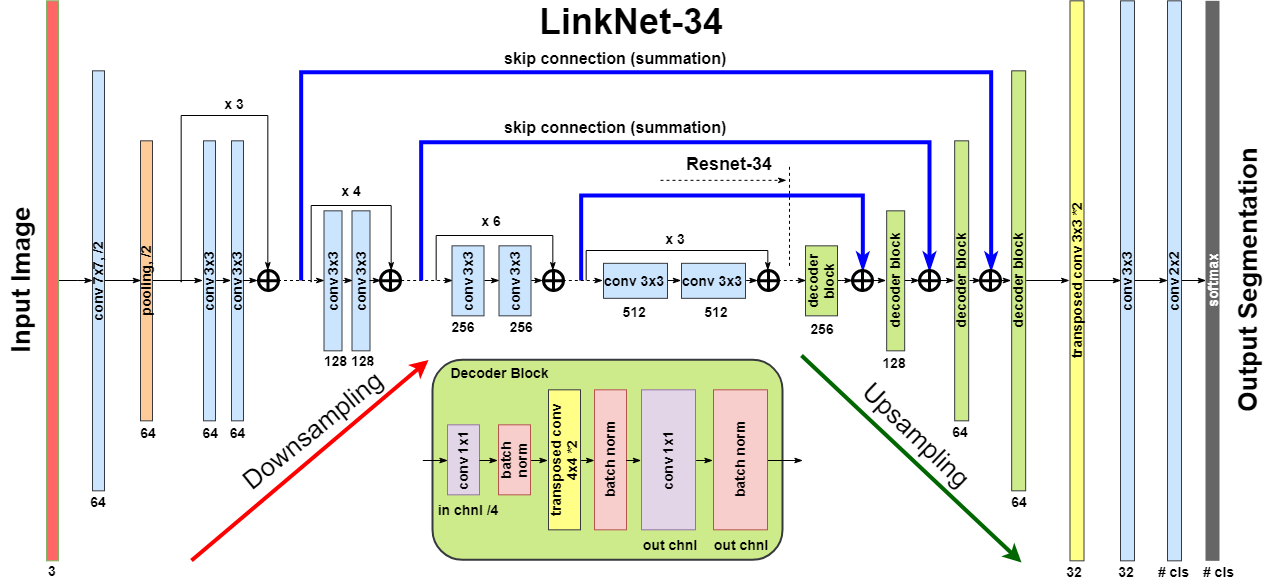}
\caption{These segmentation networks are based on encoder-decoder network of U-Net family. TernausNet uses pre-trained VGG16 network as an encoder, while LinkNet-34 uses pre-trained ResNet34. Each box corresponds to a multi-channel feature map. The number of channels is pointed below the box. The height of the box represents a feature map resolution. The blue arrows denote skip-connections where information is transmitted from the encoder to the decoder.}
\label{fig::linknet34}
\end{figure*}

In general, a U-Net-like architecture consists of a contracting path to capture context and of a symmetrically expanding path that enables precise localization (for example, see Fig. \ref{fig::linknet34}). The contracting path follows the typical architecture of a convolutional network with alternating convolution and pooling operations and progressively downsamples feature maps, increasing the number of feature maps per layer at the same time. Every step in the expansive path consists of an upsampling of the feature map followed by a convolution. Hence, the expansive branch increases the resolution of the output. In order to localize, upsampled features, the expansive path combines them with high-resolution features from the contracting path via skip-connections \cite{ronneberger2015u}. The output of the model is a pixel-by-pixel mask that shows the class of each pixel. We use slightly modified version of the original U-Net model that previously proved itself very useful for segmentation problems with limited amounts of data, for example, see \cite{iglovikov2017satellite, iglovikov2017pediatric}. Our submission to the MICCAI 2017 Endoscopic Vision SubChallenge: Robotic Instrument Segmentation \cite{miccai2017} was produced using this architecture.

As an improvement over U-Net, we use similar networks with pre-trained encoders. TernausNet \cite{iglovikov2018ternausnet} is a U-Net-like architecture that uses relatively simple pre-trained VGG11 or VGG16 \cite{simonyan2014vgg} networks as an encoder (see Fig. \ref{fig::linknet34}). VGG11 consists of seven convolutional layers, each followed by a ReLU activation function, and five max polling operations, each reducing feature map by $2$. All convolutional layers have $3\times3$ kernels. TernausNet16 has a similar structure and uses VGG16 network as an encoder (see Fig. \ref{fig::linknet34}).

In contrast, LinkNet \cite{chaurasia2017linknet} model uses an encoder based on a ResNet-type architecture \cite{he2016resnet}. In this work, we use pre-trained ResNet34, see Fig. \ref{fig::linknet34}. The encoder starts with the initial block that performs convolution with a kernel of size $7\times7$ and stride $2$. This block is followed by max-pooling with stride $2$. The later portion of the network consists of repetitive residual blocks. In every residual block, the first convolution operation is implemented with stride $2$ to provide downsampling, while the rest convolution operations use stride $1$. In addition, the decoder of the network consists of several decoder blocks that are connected with the corresponding encoder block. In this case, the transmitted block from the encoder is added to the corresponding decoder block. Each decoder block includes $1\times1$ convolution operation that reduces the number of filters by $4$, followed by batch normalization and transposed convolution to upsample the feature map.

\subsection{Training}
We use Jaccard index (Intersection Over Union) as the evaluation metric. It can be interpreted as a similarity measure between a finite number of sets. For two sets $A$ and $B$, it can be defined as following:
\begin{equation}
\label{jaccard_iou}
    J(A, B) = \frac{|A\cap B|}{|A\cup B|} = \frac{|A\cap B|}{|A|+|B|-|A\cap B|}
\end{equation}
Since an image consists of pixels, the last expression can be adapted for discrete objects in the following way:
\begin{equation}
\label{dicrjacc}
J=\frac{1}{n}\sum\limits_{i=1}^n\left(\frac{y_i\hat{y}_i}{y_{i}+\hat{y}_i-y_i\hat{y}_i}\right)
\end{equation}
where $y_i$ and $\hat{y}_i$ are a binary value (label) and a predicted probability for the pixel $i$, correspondingly.

Since image segmentation task can also be considered as a pixel classification problem, we additionally use common classification loss functions, denoted as $H$. For a binary segmentation problem $H$ is a binary cross entropy, while for a multi-class segmentation problem $H$ is a categorical cross entropy.

The final expression for the generalized loss function is obtained by combining (\ref{dicrjacc}) and $H$ as following:
\begin{equation}
\label{free_en}
L=H-\log J
\end{equation}
By minimizing this loss function, we simultaneously maximize probabilities for right pixels to be predicted and maximize the intersection $J$ between masks and corresponding predictions \cite{iglovikov2017satellite}.

As an output of a model, we obtain an image, in which each pixel value corresponds to a probability of belonging to the area of interest or a class. The size of the output image matches the input image size. For binary segmentation, we use $0.3$ as a threshold value (chosen using validation dataset) to binarize pixel probabilities. All pixel values below the specified threshold are set to $0$, while all values above the threshold are set to $255$ to produce final prediction mask. For multi-class segmentation we use similar procedure, but we set different integer numbers for each class as was noted above.

\begin{figure}[!t]
\centering
\includegraphics[width=\linewidth]{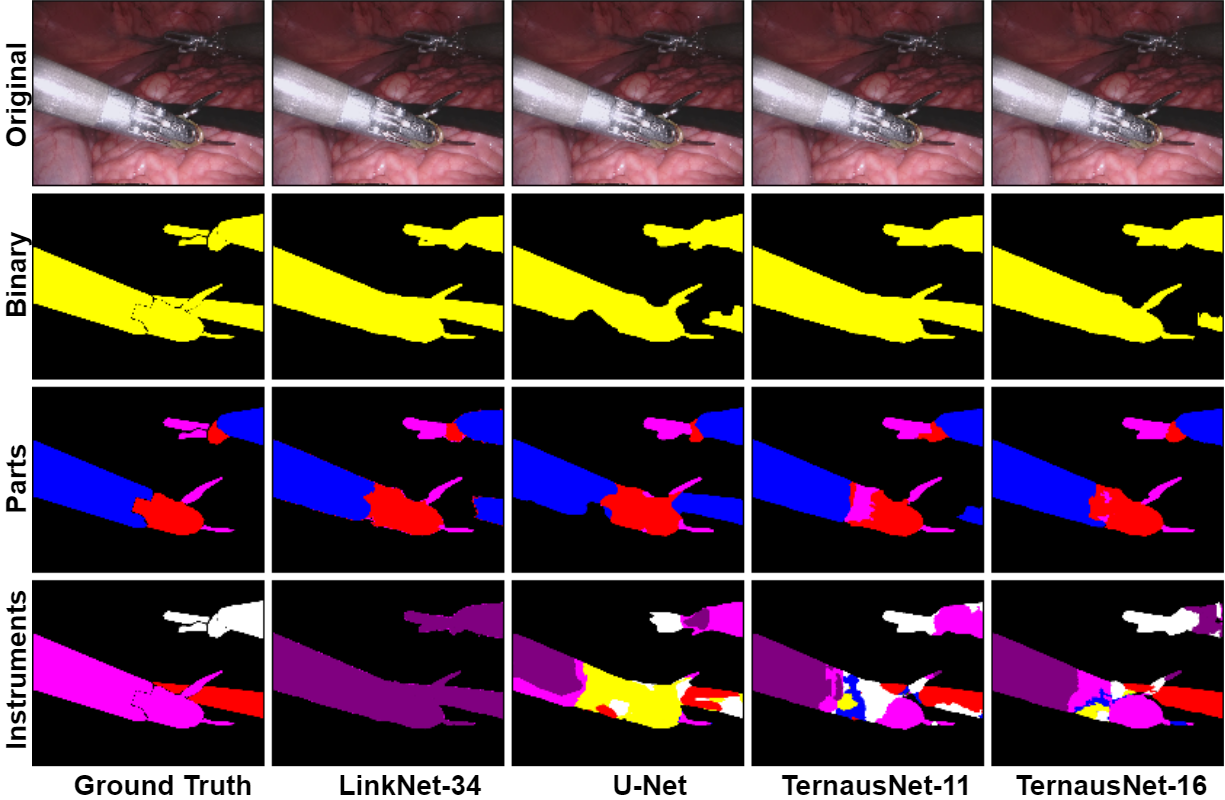}
\caption{Qualitative comparison between several  neural network architectures implemented for a binary and multi-class segmentation.}
\label{fig::grid_legend}
\end{figure}

\section{Results}
The qualitative comparison of our models both for a binary and multi-class segmentation is presented in Fig. \ref{fig::grid_legend} and Table \ref{table:segmentation}. For the binary segmentation task the best results is achieved by TernausNet-16 providing $IoU = 0.836$ and $Dice = 0.901$. These values are the best reported in the literature up to now \cite{pakhomov2017deep, garcia2017toolnet}. Next, we consider multi-class segmentation of different parts of instruments. As before, the best results reveals TernausNet-16 providing $IoU = 0.655$ and $Dice = 0.760$. For the multi-class class instrument segmentation task the results look less optimistic. In this case the best model is TernausNet-11 that achieves $IoU = 0.346$ and $Dice= 0.459$ for segmentation on 7 classes. Lower performance can be explained by the relatively small dataset size. There are 7 classes and several classes appear just few times in the training dataset.
The results suggests that this results can be  improved by increasing the dataset size for the corresponding problem.

\begin{table}[t!]
\caption{Segmentation results per task. Intersection over Union (IoU) and Dice coefficient (Dice) are in $\%$ and inference time (Time) is in $ms$.}
\label{table:segmentation}
\centering   
\begin{tabular}{|c | c | c | c | c | c | c | c | c | c|}
\hline\noalign{\smallskip}
&
\multicolumn{3}{c}{Binary segmentation} &
\multicolumn{3}{|c|}{Parts segmentation} &
\multicolumn{3}{c|}{Instrument segmentation} \\
\hline
Model & IOU & Dice & Time & IOU & Dice & Time & IOU & Dice & Time \\
\hline
U-Net & 75.44 & 84.37 & 93 & 48.41 & 60.75 & 106 & 15.80 & 23.59 & \textbf{122}\\
TernausNet-11 & 81.14 & 88.07 & 142 & 62.23 & 74.25 & 157 & \textbf{34.61} & \textbf{45.86} & 173\\
TernausNet-16 & \textbf{83.60} & \textbf{90.01} & 184 & \textbf{65.50} & \textbf{75.97} & 202 & 33.78 & 44.95 & 275\\
LinkNet-34 & 82.36 & 88.87 & \textbf{88} & 34.55 & 41.26 & \textbf{97} & 22.47 & 24.71 & 177\\

\hline
\end{tabular}
\end{table}

When compared by the inference time, LinkNet-34 is the fastest model due to the light encoder. In the case of a binary segmentation task this network takes around 90 $ms$ for $1280\times1024$ pixel image and more than twice as fast as TernausNet. The inference time was measured using one NVIDIA GTX 1080Ti GPU. A detailed comparison for the binary and multi-class tasks can be found in our GitHub repository at \url{https://github.com/ternaus/robot-surgery-segmentation}.

Suggested approach demonstrated state-of-the-art level of performance when compared to other deep learning-based solutions within to the MICCAI 2017 Endoscopic Vision SubChallenge: Robotic Instrument Segmentation \cite{miccai2017}.

\section{Conclusions}
In this paper, we describe our solution robotic instrument segmentation and demonstrate comparative analysis of various deep network models. Our approach is originally based on U-Net network architecture that we improved using state-of-the-art semantic segmentation neural networks known as LinkNet and TernausNet. Our results shows competitive performance for a binary as well as for multi-class robotic instrument segmentation. All of these networks make up end-to-end pipeline, performing efficient analysis on the whole image resolution. We believe that our methods can lay a good foundation for similar problems of real-time surgical instrument position detection. This, in turn, can be used for the tracking and pose estimation in the vicinity of surgical scenes.    


\bibliographystyle{splncs03}
\bibliography{paper}


\end{document}